\documentclass[journal,compsoc]{IEEEtran}
\ifCLASSOPTIONcompsoc
  \usepackage[nocompress]{cite}
  \usepackage[caption=false, font=normalsize, labelfont=sf, textfont=sf]{subfig}
\else
  \usepackage{cite}
  \usepackage[caption=false, font=footnotesize]{subfig}
\fi

\pdfoutput=1

\usepackage{tikz}
\usepackage{amssymb, amsmath, amsthm, amsfonts}
\usepackage{comment,verbatim}
\graphicspath{{./figures/}}
\usepackage{algorithm}
\usepackage{algorithmic}
\usepackage{adjustbox}
\usepackage{hyperref}
\usetikzlibrary{fit, calc,arrows.meta,positioning, shapes}
\usepackage{graphicx}
\usepackage{color,soul}

\newtheorem{d1}{Definition}
\usepackage{xstring}

\newcommand\copyrighttext{%
\footnotesize This work has been submitted to the IEEE for possible publication. Copyright may be transferred without notice, after which this version may no longer be accessible.}
\newcommand\copyrightnotice{%
\begin{tikzpicture}[remember picture,overlay]
\node[anchor=south,yshift=10pt] at (current page.south) {\fbox{\parbox{\dimexpr\textwidth-\fboxsep-\fboxrule\relax}{\copyrighttext}}};
\end{tikzpicture}%
}
\begin{document}
\title{NeuroKoopman Dynamic Causal Discovery}
\author{Rahmat Adesunkanmi,~\IEEEmembership{Student member,~IEEE}, Balaji Sesha Srikanth Pokuri,   \\
Ratnesh~Kumar,~\IEEEmembership{Fellow,~IEEE}
\vspace{-0.2in}
\thanks{This work was supported in part by U.S. National Science Foundation under grant NSF-CSSI-2004766 and NSF-PFI-2141084.}
\thanks{Rahmat Adesunkanmi, Balaji Sesha Srikanth Pokuri, Ph.D. students, and Ratnesh Kumar, Palmer Professor, are with the Department of Electrical and Computer Engineering, Iowa State University, Ames, IA 50010, USA (e-mail:
rahma, bspokuri, rkumar@iastate.edu.)}
\thanks{(Corresponding author: Rahmat Adesunkanmi.)}}

\maketitle
\copyrightnotice

\begin{abstract}
In many real-world applications where the system dynamics has an underlying interdependency among its variables (such as power grid, economics, neuroscience, omics networks, environmental ecosystems, and others), one is often interested in knowing whether the past values of one time series influences the future of another, known as Granger causality, and the associated underlying dynamics. This paper introduces a Koopman-inspired framework that leverages neural networks for data-driven learning of the Koopman bases, termed NeuroKoopman Dynamic Causal Discovery (NKDCD), for reliably inferring the Granger causality along with the underlying nonlinear dynamics. NKDCD employs an autoencoder architecture that lifts the nonlinear dynamics to a higher dimension using data-learned bases, where the lifted time series can be reliably modeled linearly. The lifting function, the linear Granger causality lag matrices, and the projection function (from lifted space to base space) are all represented as multilayer perceptrons and are all learned simultaneously in one go. NKDCD also utilizes sparsity-inducing penalties on the weights of the lag matrices, encouraging the model to select only the needed causal dependencies within the data. Through extensive testing on practically applicable datasets, it is shown that the NKDCD outperforms the existing nonlinear Granger causality discovery approaches. 
\end{abstract}


\section{Introduction}\label{sec:intro}
\subsection{Motivation and Related Works}
\IEEEPARstart{C}{ausality} analysis is vital in multivariate time series study for identifying potential cause and effect-informed networked dynamic relationships present in the observational data~\cite{guo2020survey}.  Granger causality (GC) corresponds to the situation in which the past of one time series helps predict the future of another~\cite{granger1969,lutkepohl2005new}. 
The notion of GC has been utilized in many fields of networked dynamics, including finance~\cite{hong2009finance},  neuroscience~\cite{reid2019advancing}, meteorology~\cite{mosedale2006granger}, economics~\cite{chiou2008economic}. In neuroscience~\cite{stokes2017study, sheikhattar2018extracting} and biology~\cite{fujita2010granger,lozano2009groupedtemporal}, for example, such analyses help determine whether a past activity in one brain region influences a later activity in another region, and to infer the underlying gene regulatory networks, respectively. 

Many traditional methods for estimating GC assume that the time series follows linear dynamics and thus use a vector autoregressive (VAR) model ~\cite{lozano2009groupedtemporal, lutkepohl2005new}, where for a vector of $n$ variables at time $t$, $x_t=(x_t(1),\hdots,x_t(n) $), the following linear regression holds:
 \begin{equation}\label{eq:VAR}
     x_t = \sum_{l=1}^L W_lx_{t-l}+e_t.
 \end{equation}
Here $W_l,l=1,\ldots, L$ are lag matrices, $L$ is the maximum lag (``memory of the system"), and $e_t$ is additive noise, typically assumed to be a zero mean white noise.
GC dependency then corresponds to the non-zero entries in $W_l,l\in[1,L]$. Authors in ~\cite{sims1972} later gave an equivalent definition of GC based on coefficients in a moving average (MA) representation (see ~\cite{chamberlain1982}). GC for an $(i,j)\in[1,n]^2$ pair can be tested statistically using an $F$-test~\cite{cromwell1994} comparing two models: an estimated model, which includes past values of both $x(i)$ and $x(j)$ versus another estimated model, that includes the past values of only $x(i)$. $x(j)$ is declared Granger causal for $x(i)$ if the $F$-test hypothesis of $x(i)$'s dependence on $x(j)$ is accepted. 

Sparsity-inducing regularizers, such as lasso~\cite{tibshirani1996regression}  or group lasso~\cite{yuan2006model}, are used to identify the minimally required number of causal dependencies~\cite{lozano2009groupedtemporal}.
These regularizers result in only a limited number of GC connections for each time series. Additionally, it is necessary to define the maximum time delay to be considered in GC analysis. To address the issue of choosing the relevant number of lags without causing overfitting, methodologies such as the hierarchical~\cite{nicholson2014hierarchical} and truncating lassos~\cite{shojaie2010discovering} have been proposed.

When data dynamics are nonlinear, imposing linear models can lead to inconsistent estimation of the true Granger causal interactions~\cite{terasvirta2010modelling, lusch2016inferring}. Model-free methods like transfer entropy~\cite{vicente2011transfer} or directed information~\cite{amblard2011directed} avoid linearity assumption, but only infer connectivity information and not the underlying dynamics~\cite{runge2012escaping}.
Deep Learning (DL) and ML methods have shown promise in learning the complex dynamics of a system. Neural networks (NNs) and their variants, such as autoregressive multilayer perceptrons~\cite{raissi2018multistep, kicsi2004river}  (MLPs) and long-short term memory networks (LSTMs)~\cite{graves2012supervised}, are capable of modeling nonlinear interactions~\cite{li2017graph}. However, these methods may lack interpretability if not carefully modeled. Accordingly, \cite{tank2021neural} introduced component-wise MLPs and LSTMs, termed cMLP and cLSTM, that support the interpretability of GC from the learned models. Search for parameters of these models is also coupled with sparsity-inducing penalties. 

In this research, we propose our approach of inferring causal structure and underlying dynamics using a  Koopman-inspired lifting of the given time series data into a higher-dimensional space where the evolution appears linear reliably; use data-driven auto-learning of the lifting and projection functions in the form of neural networks; and next infer a sparse linear regressive model in the lifted domain. The traditional approach to Koopman linear embedding employs a large number of nonlinear basis functions for approximating nonlinear systems into higher dimensional linear representations~\cite{koopman1931pnas} and for projecting the lifted linear dynamics back to the base space~\cite{williams2015jnls}. However, to select the basis functions for lifting/projection, the traditional methods rely on ad hoc predefined dictionaries of bases, such as polynomial or radial functions, which may not be optimal. Some ways to obtain approximations of the Koopman operators involve utilizing the methods of dynamic mode decomposition (DMD)~\cite{schmid2022dynamic} and the extended-DMD (EDMD)~\cite{williams2015jnls, korda2018linear}, where again, one assumes the knowledge of a predefined dictionary of basis functions. Inferring causal interactions, utilizing the Koopman embedding, is a relatively young field. Authors in~\cite{sinha2020data} infer causal interaction in a dynamical system by using transfer Perron-Frobenius (P-F) and Koopman operators, whereas~\cite{GUNJAL2023109865} utilized the setting of DMD towards GC inference. 

In contrast to the traditional predefined basis dictionary, the data-driven approach provides a potential to learn basis functions accurately, without needing any user-designated bases~\cite{GoodfellowDL}, thereby enabling a scalable auto-encoding of the basis functions. NNs have been used to learn Koopman embeddings~\cite{Wehmeyer2017arxiv, lusch2018deep}. Recently, we have also demonstrated their application to model-predictive control (MPC) of smart grids by employing auto-learned Koopman bases to accelerate MPC computation~\cite{hossain2023data}. 
In this research, we learn the GC relation and dynamics via a linear regression applied to the nonlinearly transformed time series, employing a data-driven NN-based autoencoder for the first time:
\begin{equation}\label{eq:lifted}
\varphi(x_t) =  \sum_{l=1}^L W_l\varphi(x_{t-l})+e_t,
\end{equation}
where the intrinsic nonlinear lifting map $\varphi$, the causal interactions among the lifted variables captured by lag matrices $W_l,l\in[1,L]$, as well as their projection map $\varphi^{-1}(\cdot)$, are all learned simultaneously in one go from the observed time series data. (The dimension of the auto-encoded lifted space is a hyperparameter and is chosen by exploration.)

\subsection{Our Approach and Key Contributions}
We summarize here the key contributions of our framework ``NeuroKoopman Dynamic Causal Discovery (NKDCD)" with the following key contributions:

\begin{itemize}
\item We introduce the NKDCD architecture for causal discovery and learning underlying dynamics, comprising of: 
\begin{itemize}
\item  A data-driven basis-dictionary free approach to learning the NN-based bases for linear Koopman embedding required of capturing sparse nonlinear dependency in from a higher-dimensional sparse linear model (\ref{eq:NAR});
\item A NN-based data-driven learning of sparse linear model in the lifted domain to infer GC and underlying dynamics;  
\item A data-driven learning of projection function to map down both the lifted domain time series and the estimated high-dimensional sparse linear models to the base space. 
\end{itemize}

\item The linear embedding approach is basis dictionary-free, in which the bases are auto-learned from the data, unlike the traditional approaches that manually pick basis functions that generally do not lead to optimal solutions.

\item Each element of the time series is lifted identically to its higher-dimensional representation, $x(i)\in \mathbb{R}^1 \mapsto\varphi(x(i)) \in \mathbb{R}^{1\times N },i\in[1,n]$, thereby preserving variable-wise separation at the lifted level, to ensure variable-wise disentanglement of the signals and their interdependencies in the lifted domain.

\item For preserving the structure of sparsity and interdependencies of the original variables, group-level sparsity-inducing penalties are imposed in the lifted high-dimensional setting, where all lifted variables corresponding to a single variable of the original domain are placed in the same group.

\item The autoencoder-based architecture is trained end-to-end in one go, learning directly from data the unknown basis function, the lag matrices associated with the lifted domain model, and the projection function together while minimizing the least-square prediction error and maximizing the sparsity of interdependencies. Such an end-to-end learning also eliminates the risks of numerical issues associated with iterative learning of the individual components (lifting, lag dynamics causality, and projection) \cite{hossain2023data}. 

\item For accuracy and reliability, the loss function involves multiple terms to individually account for the accuracy of each NN block's input-output relationship, making the overall autoencoder architecture robust. These correspond to the accuracy of the temporal dependencies, the GC constraints, and the lifting and projection functions.

\item Since appropriate lag selection is crucial for GC and underlying dynamics discovery, we impose additional structured group penalties that infer both the GC and the lags for each of the inferred interactions. Our formulation automatically selects a subset of the time series for each output time series that Granger causes it, regardless of the lag of the interaction.

\item Our method is validated via simulations on a variety of datasets: a finance dataset~\cite{nauta2019causal}, a nonlinear Lorenz-96 model dataset~\cite{lorenz1996predictability},  brain-imaging fMRI dataset \cite{smith2011network}, and the DREAM3 gene regulatory network benchmark dataset~\cite{prill2010towards}. Our results are able to outperform the ones reported in the literature. 
\end{itemize}

\section{Mathematical Framework}
As outlined in \cite{granger1969}, the notion of Granger causal interaction is formulated as follows: A time series  $y$ Granger causes time series $x$ if the variance of the prediction error when the history of $y$ is included is lower than the variance of the prediction error when the history of $y$ is excluded, i.e.,

\begin{align}
\sigma^2(x_t\vert \mathcal{H}_{<t})< \sigma^2( x_t |\mathcal{H}_{<t}\backslash y_{<t} ),
\end{align}
where $\mathcal{H}_{<t}$ represents the history containing all relevant information prior to time $t$, $\mathcal{H}_{<t}\backslash y_{<t}$ indicates the exclusion of $y_{<t}$ values from $\mathcal{H}_{<t}$, and $\sigma^2(\cdot)$ denotes the variance. Essentially, $y$ is causal for $x$ if past values of $y$ improve the prediction of $x$.

\subsection{Traditional VAR models for GC}\label{sec:VAR}
Considering the model of (\ref{eq:VAR}), a time series element $x(j)$  Granger-causes another time series element $x(i)$ if there exists a value at any lag $l$ for which $W_{l}(i,j) \neq 0$. GC analysis thus amounts to estimating the lag matrices from the time series observations, and determining the non-zero entries. The loss function includes the square of the prediction error together with lasso penalty~\cite{lozano2009grouped, yuan2006model} on the lag matrices to attain their sparsity: 

\begin{align} \label{eq:glasso_var}
\min_{W_{1}, \ldots, W_{L}} \sum_{t = L+1}^T & \| x_t- \sum_{l = 1}^L W_{l} x_{t - l}\|_2^2 \nonumber\\
&+ \lambda \Omega(W_{1}, \ldots, W_{L}),
\end{align}
where $\|\cdot\|_2$ denotes the $L_2$ norm, and $\lambda > 0$ is a hyper-parameter that controls the weight ascribed to the sparsity-inducing lasso term $\Omega(W_{1}, \ldots, W_{L})$ in the loss function. In our work, we explore multiple types of group lasso loss terms for inducing sparsity as described in (\ref{eq:ulgasso})-(\ref{eq:glasso}).

\subsection{Nonlinear autoregressive (NAR) model}
The VAR model assumes linear dependence, which might not be the case for a practically applicable setting, in which case, the model may not yield the GC dependencies correctly. In practice, $x_t \in \mathbb{R}^n$ can evolve according to more general nonlinear dynamics. A general element-wise nonlinear autoregressive (NAR) model is of the following form~\cite{tank2018neural}: 
\begin{align}\label{eq:NAR}
x_t(i) =f_i\bigg(x_{t-1:t-L}(1) ,\ldots,x_{t-1:t-L}(n)\bigg)+ e_t(i),
\end{align}
where $x_{t-1:t-L}(j),j\in[1,n]$ is an abbreviation to denote the past $L$ values, $x_{t-1}(j),\ldots,x_{t-L}(j)$, and $f_i$ is a function that specifies how the past lags are mapped to  $x_t(i)$. Granger non-causality of $x(j)$ for $x(i)$ corresponds to the situation when $f_i$ does not depend on $x_{t-1:t-L}(j)$.

\subsection{Proposed framework of NeuroKoopman Dynamic Causal Discovery (NKDCD)}
Suppose we have an NAR time series data ${\textbf x} = [x_1,\ldots, x_T] \in \mathbb{R}^{n\times T}$. We first lift each element of vector data at each time point to higher-dimensional intrinsic coordinates: 
\begin{align}\label{eq:Xkoop}
&X^{koop}_{t}(i) =  \varphi(x_{t}(i)) \\
&\varphi  \in \mathbb{R}^{1\times N}; X^{koop}_{t}(i) \in \mathbb{R}^{1\times N} (N>>>1 \nonumber); 
 i \in [1,n].\nonumber
\end{align}

\begin{figure*}
\centering
\includegraphics[width=.85\textwidth]{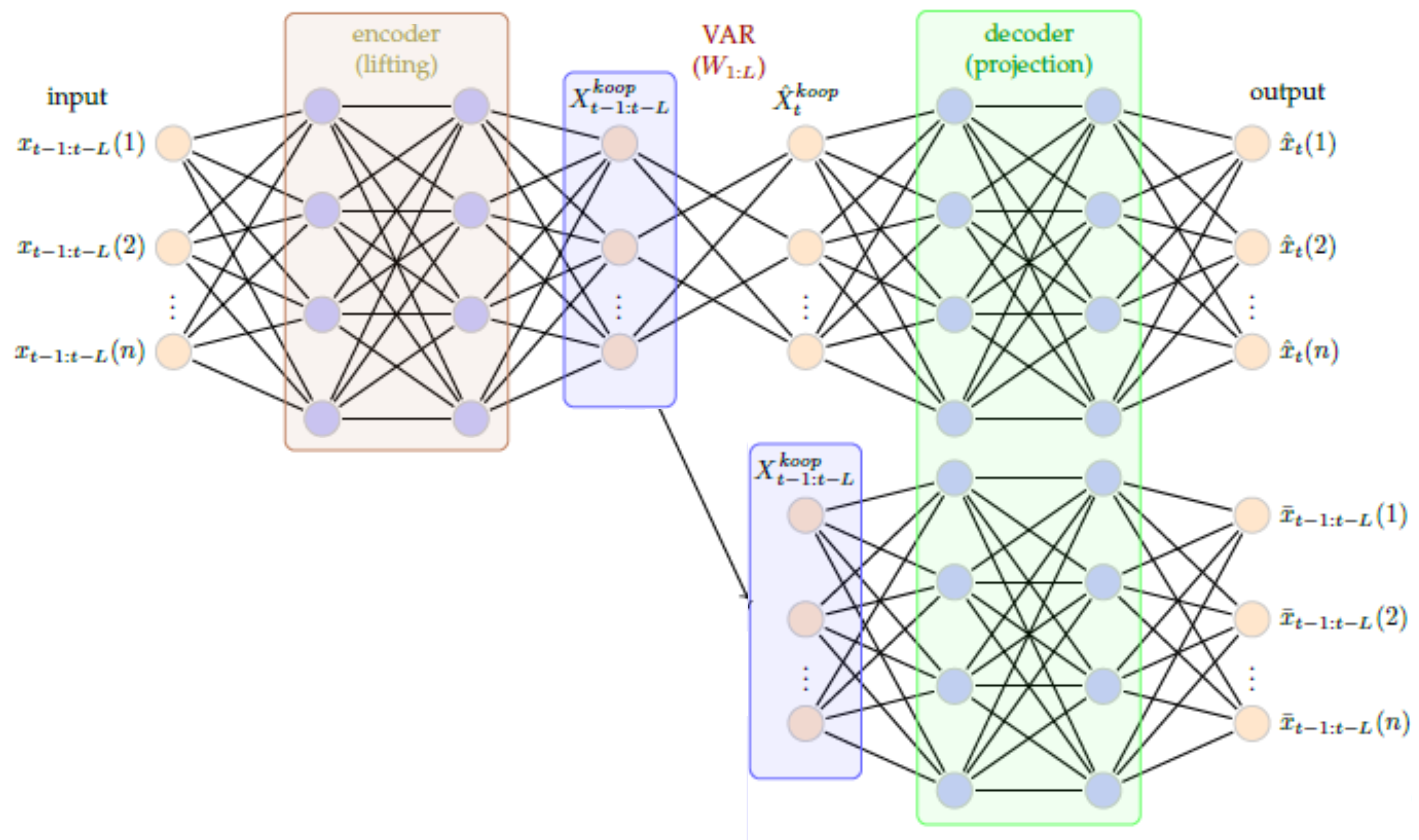}
\caption{Proposed NKDCD Architecture}
\label{fig:NA}
\end{figure*}

The lifted data is then modeled following VAR as in (\ref{eq:lifted}):
\begin{align}\label{eq:Xhatkoop}
& \widehat{X}^{koop}_t = \sum_{l=1}^L W_l X^{koop}_{t-l}\\
& W_l\in \mathbb{R}^{(n\times N)\times (n\times N)}, l\in[1,L];
 W_l(i,j) \in \mathbb{R}^{ N\times N}, i,j\in[1,n],\nonumber
\end{align}
where $W_l$'s are the lag matrices in the lifted domain, and $\widehat X^{koop}_t, t\in[L+1,T],$ is the model estimate of $X^{koop}_t$. Both $\widehat{X}^{koop}_t$ and $X^{koop}_t$ are projected element-wise using $\varphi^{-1}\in \mathbb{R}^{N\times 1}$ to the base space to yield:
\begin{align}
&\widehat{x}_t (i) = \varphi^{-1}(\hat{X}^{koop}_t(i)) \qquad t \in [L+1, T]\label{eq:smallxhat}\\
&\overline{x}_t (i)  = \varphi^{-1}(X^{koop}_t(i)) \qquad t \in [1, T].\label{eq:smallxbar}
\end{align}
Figure~\ref{fig:NA} depicts the NKDCD architecture and labels the base, lifted, and projected variables, the encoder that performs the lifting from the base space to a higher dimensional space, the decoder that performs the opposite, and the lag matrices for linear regression in the lifted domain.

\begin{d1}[NKDCD Model-based GC] \label{def:KoopNAR}
Referring to the NKDCD model (\ref{eq:Xhatkoop}), the time series $x(j)$ is Granger non-causal for time series
$x(i)$ if for all $l\in[1,L],W_l(i,j)\in\mathbb{R}^{N\times N}=0$, so the $j^{th}$ time series does not contribute to the prediction of the $i^{th}$ time series.
\end{d1}

The lifting function, $\varphi$, the projection function, $\varphi^{-1}$, and the lag matrices, $\{W_l,l\in[1,L]\}$, are learned from the observed time series data $\{x_1,\ldots,x_T\}$ in one go. To ensure learning robustness, we define a multi-term loss function, addressing the correctness of each component of the proposed NKDCD architecture.
\begin{enumerate}
\item \textit{Correctness of mapping to lifted space and its projection to base space}:  In the NKDCD framework, the dynamics evolve over the intrinsic coordinates $X^{koop} =  \varphi(x)$ in the lifted domain, whereas the inverse $\overline x =  \varphi^{-1}(X^{koop})$ recovers the base space value. For the accuracy of the lifting and projection pair, we introduce the following loss term: 
\[\sum_{t=1}^T \|x_t - \varphi^{-1}(\varphi(x_t))\|^2.\]

\item \textit{Correctness of linear VAR model in lifted space}: To discover the lag matrices that govern the GC and underlying dynamics, we learn the lag matrices $\{W_l,l\in[1,L]\}$  over the intrinsic coordinates, so that $\hat{X}^{koop}_t = \sum_{l=1}^L W_l X^{koop}_{t-l}$ holds. The accuracy of learning the lag matrices is achieved by introducing this loss term:
\[\sum_{t=L+1}^T\|\varphi(x_t) - \sum_{l=1}^L W_l \varphi(x_{t-l})\|^2.\]

\item \textit{Correctness of NAR model in base space}:  The projection operation allows us to map the linear VAR model of the lifted space to the NAR model of the base space. To achieve reconstruction accuracy of the NAR model, we introduce this additional loss term:
\[\sum_{t=L+1}^T \|x_t - \varphi^{-1}(\sum_{l=1}^L W_l \varphi(x_{t-l}))\|^2.\]

\item \textit{Correctness of NAR model with autoencoder}: The lifting and projection can be applied on the input side before comparing to the NAR output to enhance the end-to-end accuracy further, and for which we introduce this another loss term:
\[\sum_{t=L+1}^T \|\varphi^{-1}(\varphi(x_{t})) - \varphi^{-1}(\sum_{l=1}^L W_l \varphi(x_{t-l}))\|^2.\]

\item \textit{Sparsity of GC}: To have as few GC dependencies as needed, we add a fifth term to the loss function of the form $\lambda\Omega(W_1,...,W_L)$, where $\lambda>0$ is a tunable scalar hyperparameter, that penalizes having a large number of non-zero entries in $\{W_l,l\in[1,L]\}$. Since minimizing the count of non-zero entries is equivalent to minimizing the $L_0$-norm, which is discrete and hence discontinuous, one instead considers a relaxation to the $L_1$-norm. Since each $W_l(i,j),l\in[1,L]$ is a $N\times N$ matrix representing the lag-$l$ weights between the $N$ lifted components of the $i$th time series element and the $N$ lifted components of the $j$th time series element, all such $N\times N$ elements of $W_l(i,j)$ for each $(i,j)\in[1,n]^2,l\in[1,L]$ are grouped together, thereby the sparsity penalty becomes an $L_1$-norm over groups, termed group lasso.

\hspace*{10pt} In addition, since all elements $W_l(i,j),l\in[1,L]$ must be zero to ensure that $x(i)$ is not GC dependent on $x(j)$, a grouping across the lags can be further employed. Grouping all the lags into a single group treats them equally, resulting in a ``uniform-lag group lasso (ulg-lasso)": \begin{align}\label{eq:ulgasso}
\Omega(W_1,...,W_L)\!=\!\!
\sum_{ij}\! \|W_{1}(i,j), \ldots, W_{L}(i,j)\|_2.
\end{align}
\hspace*{10pt} The ulg-lasso of (\ref{eq:ulgasso}), however, runs the risk of treating the distant lag terms that may be absent in the ground truth the same as the near ones that are actually present in the ground truth. So a ``hierarchical-lag group lasso (hlg-lasso)" that discounts more distant lags more severely is instead used sometimes:
\begin{align}\label{eq:hlasso}
\Omega(W_1,...,W_L)=\sum_{ij} \sum_{l = 1}^L \|W_{l}(i,j), \ldots, W_{L}(i,j)\|_2.
\end{align}
\hspace*{10pt} Finally, the most general case is when all the lags, near or distant, are treated independently, resulting in ``independent-lag group lasso (ilg-lasso)":
\begin{align}\label{eq:glasso}
\Omega(W_1,...,W_L)=\sum_{ijl} \|W_{l}(i,j)\|_2.
\end{align}
Note as discussed above, the grouping in (\ref{eq:glasso}) corresponds to putting all $N\times N$ elements of $W_l(,i,j)$ into its own group for each $(i,j)\in[1,n]^2,l\in[1,L]$. Such grouping is also applicable to ulg- and hlg- lassos.
\end{enumerate}

In summary, the overall loss function is defined as:
\begin{align}\label{eq:loss}
\cal{J}&=\sum_{t=1}^T \|x_t - \bar x_t\|^2+\sum_{t=L+1}^T \|X^{koop}_t - \hat{X}^{koop}_t\|^2
 \nonumber\\
&+ \sum_{t=L+1}^T \|x_t - \hat x_t\|^2+ \sum_{t=L+1}^T \|\bar x_t - \hat x_t\|^2 
 \nonumber\\
&+\lambda\Omega(W_1,...,W_L),
\end{align}
where $\Omega(\cdot)$ takes the one of the forms in (\ref{eq:ulgasso})-(\ref{eq:glasso}). For notational convenience, we denote the first four terms in $\mathcal J$ as ${\cal J}_1$ (which involves four $L_2$-norm terms, each of which is differentiable and so allows gradient descent for loss minimization) and denote the last term in $\mathcal J$ as ${\cal J}_2$, which being group lasso, involves inter-group $L_1$-norm and intra-group $L_2$-norm, that is not differentiable, requiring proximal-gradient descent for ${\mathcal J}_2$-loss minimization (in contrast, the ${\mathcal J}_1$-loss minimization uses the usual gradient descent).

In Section~\ref{sec:lasso_compare}, we employ all three instances of group lasso for comparison and demonstrate how ilg-lasso outperforms hlg-lasso, which in turn surpassed ulg-lasso in performance.

\subsection{Hyperparameters of NKDCD and loss function}
In our autoencoder model of Figure~\ref{fig:NA}, each time series is lifted to $N$ dimensions, which is one of the hyperparameters. The maximum amount of lag is $L$, which is another hyperparameter. We have chosen the lifting and projection functions to have $2$ hidden layers, where when the data has nonlinear dynamics, each hidden layer uses the leaky rectified linear unit (leakyReLU) as the activation function: $max\{0.1y, y\}$ and no activation function when the data has linear dynamics.    
The lifting hidden layer weight sizes are parameterized by $h$ and are respectively, $1\times \frac{h}{2} $, $\frac{h}{2}\times h$, and $h\times N$, while the projection hidden layer weight sizes are respectively, $N\times h$, $h\times \frac{h}{2} $, and $\frac{h}{2}\times 1$. The hyperparameter of the loss function is the sparsity weight factor $\lambda$. During the training of NKDCD, the learning rate (also called step-size) $\tau$ is yet another hyperparameter, and so is the threshold $\epsilon>0$ used to view a weight parameter below the threshold as zero. 

\subsection{NKDCD optimization: (Proximal) gradient descent}
Recall the decomposition ${\cal J}(={\cal J}_1+{\cal J}_2)$ of the loss function,
in which the ${\cal J}_2$ portion of the loss function is only a function of the lag matrices $\{W_l,l\in[1,L]\}$, and it does not depend on the lifting and projection functions $\varphi, \varphi^{-1}$. As a result, the gradient of loss ${\cal J}$ with respect to $\varphi$, denoted $\Delta_\varphi\!({\cal J})$, satisfies: $\Delta_\varphi\!({\cal J})=\Delta_\varphi\!({\cal J}_1)$, and similarly, $\Delta_{\varphi^{-1}}\!({\cal J})=\Delta_{\varphi^{-1}}\!({\cal L}_1)$. In contrast, the gradient of $\cal J$ vs. ${\cal J}_1$ with respect to the lag matrices are different from each other: $\Delta_{W_l}\!( {\cal J})\neq\Delta_{W_l}\!({\cal J}_1),~l\in[1,L]$, because $\Delta_{W_l}\!({\cal J}_2)\neq 0, l\in[1,L]$ (whereas in contrast, $\Delta_{\varphi}\!({\cal J}_2)=\Delta_{\varphi^{-1}}\!({\cal J}_2)=0$).

Accordingly, in the $m$th iteration ($m\geq 0$) of update, the lifting and projection functions are updated using gradient descent with respect to only the respective ${\cal J}_1$ portions of the gradients, applied to the corresponding values from the $m$th iteration, as follows:
\begin{align}
   \varphi^{(m+1)}&= \varphi^{(m)} - \tau \Delta_{\varphi}\!({\cal J}_1)\!\left|_{W_{1:L}^{(m)},\varphi^{(m)},(\varphi^{-1})^{(m)}}\right.,\label{eq:phibackprop}\\
     (\varphi^{-1})^{(m+1)}&= (\varphi^{-1})^{(m)} - \tau \Delta_{\varphi^{-1}}\!({\cal J}_1)\!\left|_{W_{1:L}^{(m)},\varphi^{(m)},(\varphi^{-1})^{(m)}},\right.\label{eq:invphibackprop} 
\end{align}
in which $\tau>0$ is the step size of the update (also called learning rate), and $(\cdot)^{(m)}$ denotes the value of $(\cdot)$ at an $m~\geq~0$ iteration of the update. In contrast, the update for the lag matrices follows the following computation in the case of ulg-lasso, involving gradient descent for the ${\mathcal J}_1$ part of the loss (that is differentiable) and subsequently proximal gradient descent for the ${\mathcal J}_2$ part of the loss (that is not differentiable):
\begin{align}
&\widetilde W^{(m)}_l:=\!W_l^{(m)} - \tau \Delta_{W_l}\!({\cal J}_1)\!\left|_{W_{1:L}^{(m)},\varphi^{(m)},(\varphi^{-1})^{(m)}}\right.\label{eq:lagbackprop_intermediate}\\
&W_l^{(m+1)}=\text{prox}_{\tau\lambda\Omega}\big(\widetilde W^{(m)}_l\big),\label{eq:lagbackprop}\\
&\text{prox}_{\tau\lambda\Omega}\big(\!\cdot\!\big)(i,\!j)\!=\!\bigg(\!\!1\!-\!\frac{\tau\lambda}{\vert\vert W_1^{(m)}(i,\!j)\!,\!\ldots\!,\!W_L^{(m)}(i,\!j)\vert\vert _2}\!\bigg)_{\!+}\!\!\!\big(\!\cdot\!\big)(i,\!j), \label{eq:proxupdate}
\end{align}
where the notation $\big(\!\cdot\!\big)_+:=\max\{0,\big(\!\cdot\!\big)\}$, and (\ref{eq:proxupdate}) computes the proximal gradient~\cite{parikh2014proximal} for the case of ulg -group lasso. In the case of the hlg-lasso, the proximal gradient is computed iterating over a total of $L$ rounds, where in the $l^{th}$ round $(l\geq 1)$, the subset $\big\{W_l^{(m,l)},\ldots, W_L^{(m,l)}\big\}$ of  round-$l$ values are updated using the corresponding round-$(l-1)$ ones:
\begin{align}
&\text{prox}_{\tau\lambda\Omega}\big(\!\cdot\!\big)(i,\!j)\nonumber\\
&=\!\bigg(\!\!1\!-\!\frac{\tau\lambda}{\vert\vert W_l^{(m,l-1)}(i,\!j)\!,\!\ldots\!,\!W_L^{(m,l-1)}(i,\!j)\vert\vert _2}\!\bigg)_{\!+}\!\!\!\big(\!\cdot\!\big)(i,\!j), \label{eq:proxupdate_hl}
\end{align}
where for the round-1 update $(l=1)$, the values used are:
\[W_l^{(m,0)}=\widetilde W^{(m)}_l,l\in[1,L].\]
Finally, the proximal gradient for ilg-lasso uses:
\begin{align}
&\text{prox}_{\tau\lambda\Omega}\big(\!\cdot\!\big)(i,\!j)\!=\!\bigg(\!\!1\!-\!\frac{\tau\lambda}{\vert\vert W_l^{(m)}(i,\!j)\vert\vert _2}\!\bigg)_{\!+}\!\!\!\big(\!\cdot\!\big)(i,\!j). \label{eq:proxupdate_il}
\end{align}

It can be seen that $W_l^{(m+1)},l\in[1,L],m\geq 0$ is computed from $W_l^{(m)}$ by first applying the gradient descent to obtain an intermediate value: $\widetilde W^{(m)}_l=W_l^{(m)} - \tau \Delta_{W_l}\!({\cal J}_1)\!\left|_{W_{1:L}^{(m)},\varphi^{(m)},(\varphi^{-1})^{(m)}}\right.$, accounting for only the ${\cal J}_1$ portion of the loss (ignoring the ${\cal J}_2$ portion of the loss for the time being), and next, the value nearest to this intermediate value that also accounts for the ${\cal J}_2$ portion of the loss is found by applying the proximal gradient, $\text{prox}_{\tau\lambda\Omega}\big(\!\cdot\!\big)$, which for the ulg-lasso as well as ilg-lasso is given by a single round computation, employing Eqs.~(\ref{eq:proxupdate}) and (\ref{eq:proxupdate_il}), respectively, and for the hlg-lasso it comprises of $L$-rounds of computations, where for each $l\in[1,L]$, the round-$l$ computation is given by Eq.~(\ref{eq:proxupdate_hl}). 

Considering the ulg-lasso penalty of (\ref{eq:ulgasso}) over the lag matrices $W_l(i,j),l\in[1,L]$, it shrinks each weight $W_l(i,j)$ equally, using (\ref{eq:proxupdate}). The hlg-lasso proximal gradient in (\ref{eq:hlasso}), however,  shrinks more those lag matrices that are further in the distant past, by using  (\ref{eq:proxupdate_hl}), which for each $l\geq 1$, applies $l$-rounds of shrinkage operations (so $W_1(i,j)$ undergoes only one round of shrinkage, whereas $W_L(i,j)$ witnesses $L$ rounds of shrinkage). Finally, the shrinkage operation in the case of ilg-lasso (\ref{eq:glasso}) is independent for each lag using (\ref{eq:proxupdate_il}) and is also applied in a single round (like ulg-lasso).
The resulting update steps are summarized in Algorithm~\ref{NKDCD}.

\begin{algorithm}\caption{NKDCD}\label{NKDCD}
\begin{algorithmic}
\REQUIRE time series data, ${\textbf x} \in \mathbb{R}^{n\times T}$,   $\lambda >0$, $1\geq\tau > 0$, $N>>>1$, $L>0$,\\
Initialize: $m=0,W_l^{(m)}$, $\varphi^{(m)}$, $(\varphi^{-1})^{(m)}$
\WHILE{not converged}
\STATE compute $\{X^{koop}\}$ as in Eq.~\ref{eq:Xkoop}
\STATE compute $\{\hat{X}^{koop}\}$ as in Eq.~\ref{eq:Xhatkoop}
\STATE compute $\{\hat{x}\}$ as in Eq.~\ref{eq:smallxhat}
\STATE compute $\{\bar{x}\}$ as in Eq.~\ref{eq:smallxbar}
\STATE compute loss ${\cal T}$ as in Eq.~\ref{eq:loss}
\STATE Update $\varphi^{(m)},(\varphi^{-1})^{(m)},\{W_l^{(m)},l~\in~[1,L]\}$~using~(\ref{eq:phibackprop})~-~(\ref{eq:proxupdate_il}) to get $\varphi^{(m+1)},(\varphi^{-1})^{(m+1)},\{W_l^{(m+1)},l\in[1,L]\}$
\STATE  $m = m+1$
\ENDWHILE\\
\RETURN  $\varphi,\varphi^{-1},\{W_l,l\in[1,L]\}$ 
 \end{algorithmic}
\end{algorithm}

\section{Comparing Group Lasso Penalties}\label{sec:lasso_compare}  
\begin{figure}[!htb]
    \centering
    \subfloat[\footnotesize Ground truth causal graph\label{varg}]{
    \includegraphics[width=0.45\linewidth]{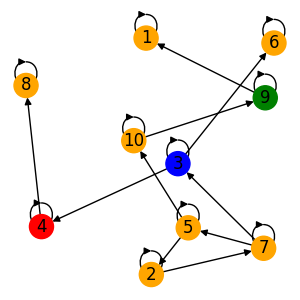}}
    \hfill
        \subfloat[\footnotesize Observed time series\label{vart}]{
       \includegraphics[width=0.45\linewidth]{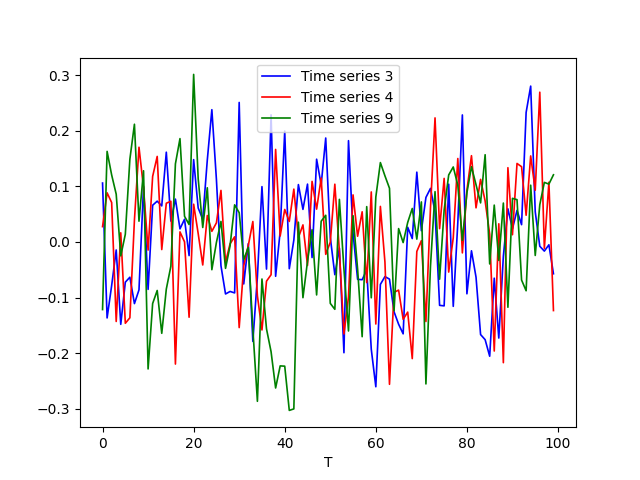}} 
    \caption{VAR(3) data}
    \label{fig:ogvar}
    \end{figure}

To qualitatively visualize the sparsity pattern recovered under the three lasso penalties, we apply our framework to data generated from a sparse VAR(3) model with $n = 10$, presented in \cite{tank2021neural}. Here, each time series $i$ depends on itself together with one other randomly selected time series among the remaining $n-1$ ones, and when time series $i$ depends on time series $j$,  $W_l(i,j)=0.1$ for $l = 1,2,3$, while all the other lag entries are zero. Thus, this setting corresponds to the ground truth of $L=3$. The ground truth causal graph and three selected time series are depicted in Figure~\ref{fig:ogvar}. Time series $3$ and $4$ exhibit a causal relationship, while neither is causally linked to the time series $9$. We collect $T=1000$ sample points per time series for analysis.

\begin{figure}[!htb]
    \centering
        \subfloat[\footnotesize Uniform-lag group lasso estimated lag matrices\label{varoggl}]{\hspace*{-.1in}
    \includegraphics[width=\linewidth]{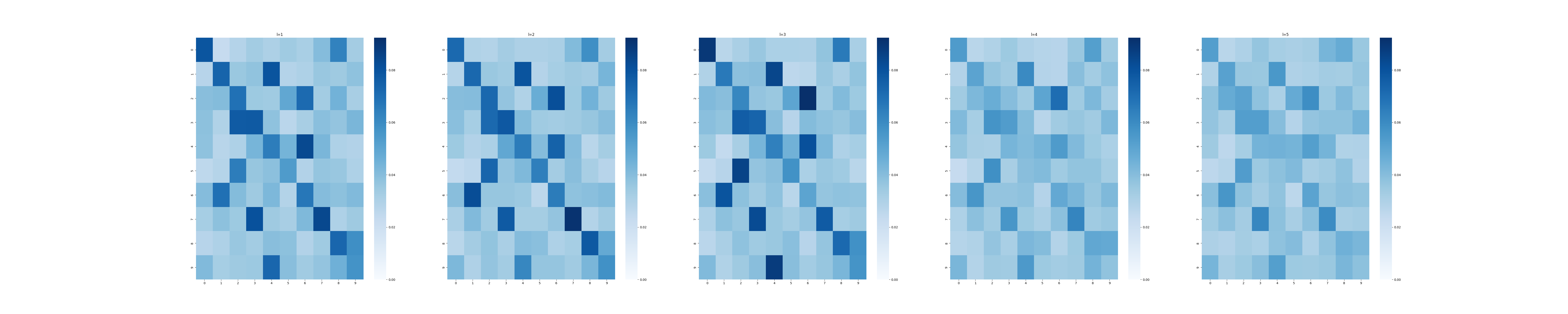}}
    \hfill
    \subfloat[\footnotesize Hierarchical-lag group lasso estimated lag matrices\label{varhl}]{\hspace*{-.1in}
       \includegraphics[width=\linewidth]{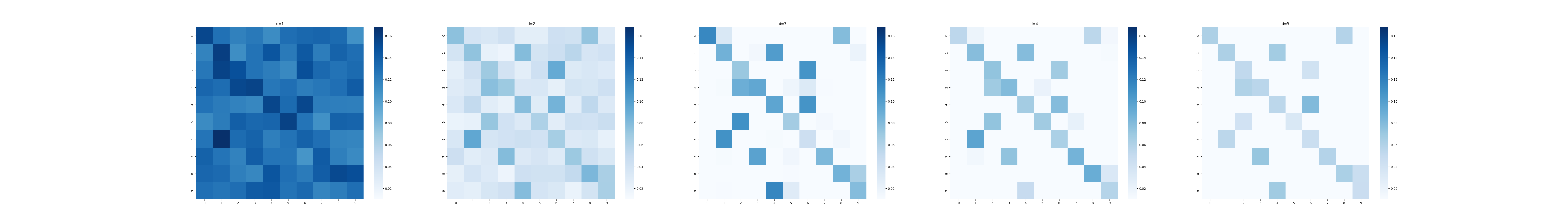}} 
     \hfill   
    \subfloat[\footnotesize Independent-lag group lasso estimated lag matrices\label{vargl}]{\hspace*{-.1in}
    \includegraphics[width=\linewidth]{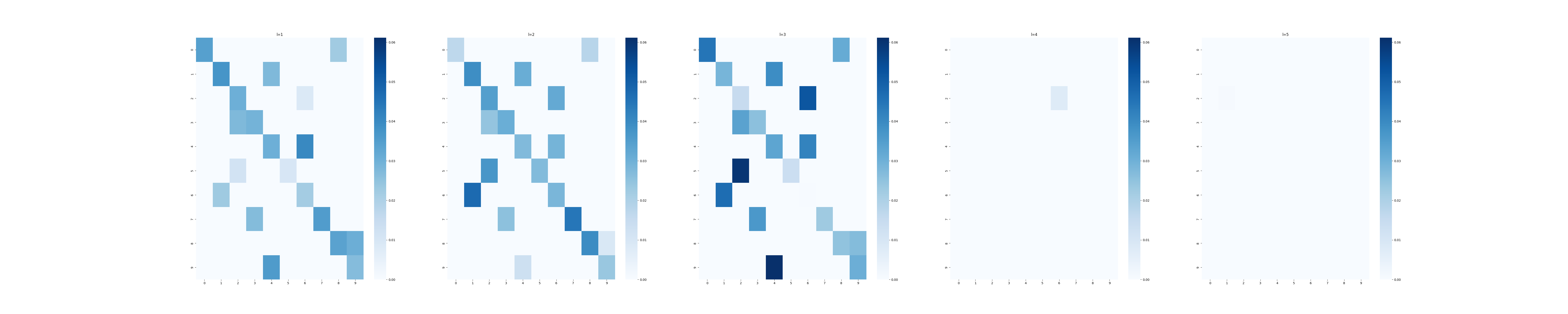}}
    \caption{ Estimated 5-step lag matrices}
    \label{fig:varest}
    \end{figure}

Because the data already follows linear dynamics, the nonlinear activation in each hidden layer was omitted. For causal model discovery, we set the learning rate $\tau = 5\times 10^{-2}$, the number of hidden layers in lifting and projection networks as 2 with $h=4$, the scaling-up factor of each dimension during lifting as $N=10$, batch size $=500$, and the upper bound to lag $L=5$. $\lambda = 2\times 10^{-2}$  was chosen for the ulg- and ilg-lassos as an experimentally determined based penalty factor, while $\lambda = 10^{-4}$ was used for the hlg-lasso, which again is the corresponding best value determined experimentally. The entries of the lag matrices are shown as heat-maps in Figure~\ref{fig:varest}, where each $(i,j)$ entry is the norm of the $N\times N$ entries in the lifted-version $W_l(i,j)$ (all of which belong to a single group).  

Taking a closer look at the captured blocks, i.e., for which $ \|W_l(i,j)\| > \epsilon$ for a suitable threshold value of epsilon (chosen experimentally), it can be seen in the cases of ilg- and hlg-lassos that if $ \|W_l(i,j)\|$ is shrunk to zero for some $l$, then $ \|W_{l'}(i,j)\|,l'>l$ is also shrunk to zero. ilg- and hlg-lassos have different stopping epochs: $2500$ vs $6200$, which implies ilg-lasso converges faster than hlg-lasso. On the other hand, while ulg-lasso converges just as fast as ilg-lasso: $2500$ iterations, it cannot accurately match the right value of $L$ (as noted above, it runs the risk of treating the distant lag terms that may be absent in the ground truth the same as the near ones that are actually present in the ground truth). Based on these comparative observations of the three group lassos, moving forward, we have chosen to apply the NKDCD computations only for ilg- and hlg-lasso (both of which outperform ulg-lasso).

\section{Results for Practical Applications}
Our study demonstrates the performance of our novel NKDCD approach in inferring Granger causality (GC) and underlying dynamics from both synthetic and practically applicable datasets. We conduct experiments on financial data~\cite{nauta2019causal}, Lorenz-96 model data~\cite{lorenz1996predictability}, fMRI brain-imaging model data~\cite{nauta2019causal}, and DREAM3 genetic data~\cite{prill2010towards}, sourced from various existing studies.

The results indicate that our NKDCD approach effectively discovers the underlying GC dependencies and nonlinear dynamics across various datasets. Key components of our model include a lifting encoder comprising multilayer perceptron (MLP) with $2$ hidden layers, GC weight matrices, and a projection decoder, another MLP with $2$ hidden layers. Each node in the encoder and decoder MLP has an activation function of a leakyReLU. We explore different hyperparameters, such as the learning rate ($\tau$) varied over $[5 \times 10^{-4},5 \times 10^{-3}$], and the sparsity weight ($\lambda$) varied over $[10^{-2},5 \times 10^{-2},3\times 10^{-1}]$. See Table~\ref{tab:hyper} for a complete list of the hyperparameters chosen experimentally that we use in analyzing the application datasets. We set a convergence criterion as the average loss per time series being less than the threshold of $0.9$ and is no longer decreasing. The performance metrics are based on the area under the ROC curve (AUROC) and the area under the Precision-Recall curve (AUPR). We compare our results against linear VARs with hlg- and ilg-lasso for sparsity as formulated in Eqs.~(\ref{eq:glasso_var}), (\ref{eq:hlasso}), and (\ref{eq:glasso}) for baseline comparison, as well as the state-of-art results available in literature, including component-wise MLP (cMLP)/LSTM (cLSTM) as reported in~\cite{tank2021neural}, to demonstrate the gain resulting from the proposed NKDCD approach over the state-of-art. In our results, the weights corresponding to lag matrices, if they are small compared to a threshold $\epsilon>0$ (decided case-by-case for optimal result), are viewed zero to infer GC. Our NKDCD approach shows promising results, outperforming the ones reported in the literature.

\begin{table}[hbt!] 
\centering
 \caption{Hyperparameters used for NKDCD Analysis}
 \label{tab:hyper}
 \resizebox{\columnwidth}{!}{
\begin{tabular}{|| c c c c c c||} 
 \hline 
 & Finance &Lorenz-96&fMRI  &Dream3  &Dream3  \\
  & && &(EC1,EC2,Y1) &(Y2,Y3) \\\hline \hline
$\lambda$ &$.01$ &$.05$ & $.05$&$.01$ &$.3$ \\
 $\tau $&$.005$& $.0005$ &$.0005$&$.0005$&$.005$ \\ 
 $L$ &$3$ &$5$ &$3$& $2$&$2$\\
 $N$ &$15$ &$15/45$ &$10$ &$5$&$10$\\ 
 $h$ &$4$ & $16$&$8$& $8$&$8$\\ 
 batch &$1024$&$500$&$500$&$966$&$966$\\ \hline
 \end{tabular}}
\end{table}

\subsection{Finance Data}
\begin{figure}[!htb]\label{fig:ogfinance}
    \centering
    \subfloat[\footnotesize Ground truth causal graph \label{fig:fina}]{%
    \includegraphics[width=0.45\linewidth]{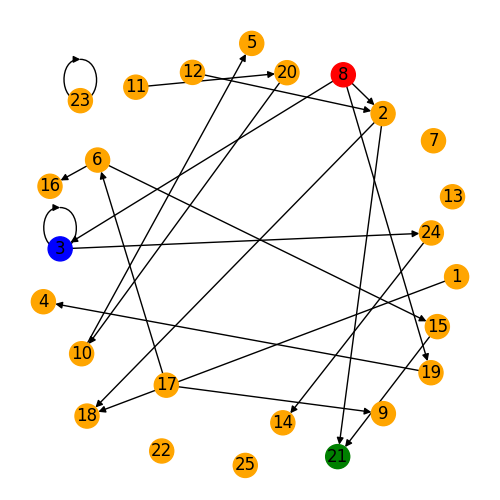}}
    \hfill
        \subfloat[\footnotesize Observed time series\label{fig:finb}]{%
       \includegraphics[width=0.55\linewidth]{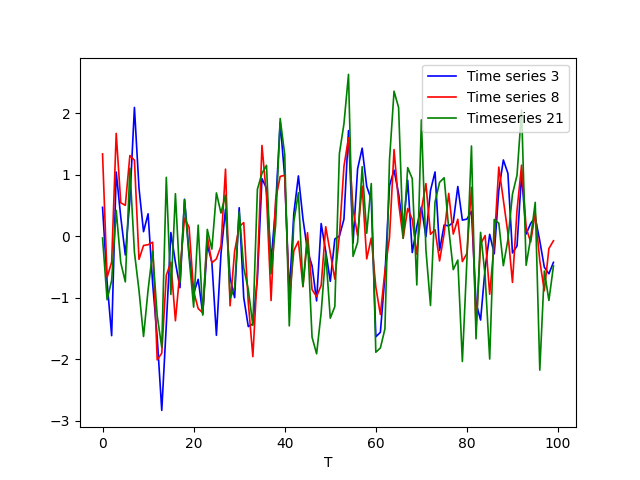}} 
    \caption{Finance data }
    \end{figure}

The simulated finance dataset utilized in our study originates from research reported in~\cite{nauta2019causal} and is publicly available on GitHub. This dataset is fitted to the Fama-French Three-Factors Model~\cite{fama1992cross}, a framework that characterizes stock returns against three factors: volatility, size, and value. For our experiments, we specifically utilize dataset 20-1$A$, which comprises a $4000$-day observation period across $25$ stocks. In ~\cite{nauta2019causal}, the causal relationships among the time series present in this dataset evolve according to linear dynamics, and $L$ is within 1-3. The ground truth interdependency is provided in Figure~\ref{fig:fina} and a sample of three selected time series elements is depicted in~\ref{fig:finb}, where the causal relationship between time series $3$ and $8$ is evident. At the same time, neither has a causal connection with the time series $21$. In this analysis, we set the maximum lag ($L$) to $3$, and since the data already follows linear dynamics, the nonlinear activation (namely, leakyReLU) is omitted within the hidden layers.

\begin{table}[ht]\caption{Finance Data Results}\label{tab:fin}
\centering
\begin{tabular}{||c| c c||} 
 \hline
 Dataset&\multicolumn{2}{|c||}{20-1$A$}  \\
  \hline
 Metrics &AUROC&AUPR\\ \hline \hline
NKDCD+Eq.~\ref{eq:glasso} & $99.9\pm 0.1$ &$98.0\pm1.0$  \\ 
NKDCD+Eq.~\ref{eq:hlasso}  & $99.1\pm 0.0$&$ 90.9\pm 0.1$  \\
VAR+Eq.~\ref{eq:ulgasso}  &$99.9\pm 0.1$ & $97.6\pm 1.0$  \\
VAR+Eq.~\ref{eq:hlasso}  &$99.9\pm 0.1$ & $97.6\pm 1.0$  \\
 \hline
 \end{tabular}  
\end{table}

The performance metrics are reported as the mean across five initializations with a $95\%$ confidence interval. Results from our experiments on this finance dataset are summarized in Table~\ref{tab:fin}. The AUROC and AUPR for the VAR model with the ulg- and hlg-lasso penalties are both $99.9\pm 0.1$ and $97.6\pm 1.0$, respectively. Our method performs just as well as the traditional VAR by giving AUROC and AUPR of $99.9\pm 0.1$ and $98.0\pm 1.0$ for the ilg-lasso penalty and $ 99.1\pm0.0$ and $90.9\pm 0.1$ respectively for the hlg-lasso counterpart,  showing that Koopman lifting preserves the causal relationship among the time series elements. These results demonstrate the effectiveness of our approach in uncovering the causal relationships and the underlying dynamics within the financial data having the 3-Factors linear model.    

\subsection{Lorenz-96 Model with Nonlinear Dynamics}
\begin{figure}[!htb]
    \centering
    \subfloat[\footnotesize{Ground truth causal graph} \label{lgraph}]{\includegraphics[width=0.45\linewidth]{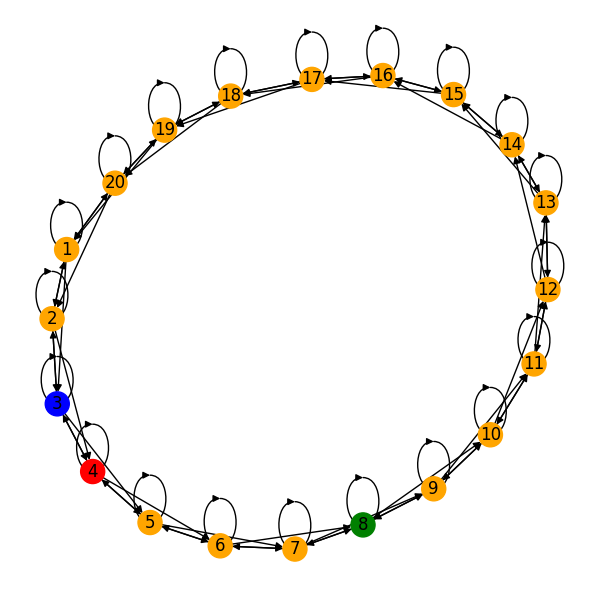}}
    \hfill
    \subfloat[\footnotesize Observed time series (selected) \label{lts}]{\includegraphics[width=0.55\linewidth]{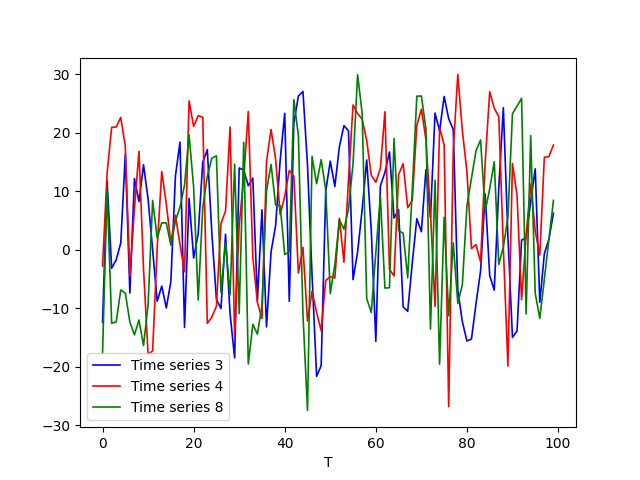}}
    \caption{Lorenz-96 model data}
    \label{fig:ogLorenz96}
\end{figure}

The Lorenz-96 model, introduced in \cite{lorenz1996predictability}, serves as a valuable tool for studying essential aspects of chaos theory, predictability, and the behavior of complex nonlinear interdependent dynamics. It provides a simplified yet insightful representation of the atmospheric dynamics, making it widely used in atmospheric science and related fields. The continuous dynamics in a Lorenz-96 model is defined for $i=1, \ldots, n, n\geq 4$ as follows:
\begin{align}\label{eq:lorenz96}
   \frac{d x_t(i)}{dt} = \bigg(  x_t(i+1) - x_t(i-2) \bigg)  x_t(i-1) -   x_t(i) +F,
\end{align}   
along with the following boundary correspondences:
\begin{align*}
    x_t(-1) =  x_t(n-1); x_t(0) =  x_t(n); x_t(n+1) =  x_t(1),
\end{align*}
\noindent where $F$ is a forcing constant that determines the level of chaos, and $n$ is the dimension. For comparison with recent work, we utilized publicly available data of a Lorenz-96 model, described by Eq.~\ref{eq:lorenz96} on GitHub, supplied by~\cite{tank2021neural}. For our analysis, we chose $n=20$ and a sampling rate of $0.1$ for obtaining a discrete-time time series. The ground truth causal graph and three selected time series for  $F = 40$ are depicted in Figure~\ref{fig:ogLorenz96}. Notably, time series $3$ and $4$ exhibit a causal relationship, while neither is causally linked to time series $8$.

 \begin{table*}[t]
\centering
  \caption{Lorenz-96 Data AUROC Results as a function of Forcing Constant $F$ and time series Length $T$} \label{tab:lorenz96}  
\begin{tabular}{||c| c c c| c c c||} 
 \hline
 &\multicolumn{3}{|c|}{$F=10$} & \multicolumn{3}{|c|}{$F=40$}  \\
  \hline
 $T$& $250 $& $500 $& $1000$  & $250 $& $500$ & $1000$ \\ [0.5ex] 
 \hline\hline
NKDCD+Eq.~\ref{eq:glasso} &$94.1\pm 0.7$&$98.5\pm 0.2$&$99.8\pm 0.1$ & $80.8\pm 0.1$ &  $92.0\pm 1.1$ &$98.7\pm 0.1$ \\ 
NKDCD+Eq.~\ref{eq:hlasso}  & $94.1\pm 0.6$ & $98.5\pm 0.1$ & $99.7\pm 0.1$  &$82.6\pm 0.3$ & $90.8\pm0.7$ & $96.7\pm 0.2$\\
cMLP& $86.6 \pm 0.2$&$96.6\pm 0.2$&$98.4\pm 0.1$ &$84.0\pm 0.5$ & $ 89.6  \pm0.2$ &$95.5  \pm 0.3$\\
cLSTM &$81.3  \pm 0.9$ & $93.4 \pm 0.7$&$ 96.0  \pm 0.1$&$ 75.1 \pm 0.9 $&$87.8 \pm 0.4$& $94.4 \pm0.5$\\
VAR+Eq.~\ref{eq:ulgasso}&$86.1\pm 0.6$ & $86.2\pm 0.2$  &$93.0\pm 0.4$ &$70.0\pm 0.0$ & $80.6\pm 0.0$&$84.7\pm0.0$\\
VAR+Eq.~\ref{eq:hlasso}   &$86.1\pm 0.6$ &$86.2\pm 0.2$  & $93.0\pm 0.4$& $70.4\pm 0.0$& $81.1\pm 0.0$&$84.7\pm0.0$\\
 \hline
 \end{tabular}
\end{table*}

We conduct simulations for $F=10,40$ and $T=250,500,100$, as specified in Table~\ref{tab:lorenz96}, where our objective is to compare the performance of the NKDCD approach against other state-of-art methods in literature, including component-wise MLP (cMLP)/LSTM (cLSTM) as reported in~\cite{tank2021neural}, and the standard linear VAR method. The performance metrics are reported as the mean across five initializations with a $95\%$ confidence interval.

The results, summarized in Table~\ref{tab:lorenz96}, reveal a tradeoff between accuracy and length of observation as $F$ increases. Despite challenges posed by shorter data lengths and higher chaos levels, NKDCD demonstrates superior performance over alternative methods in most scenarios, except for a single scenario when the data length is small ($T=250$), and chaos is greater ($F=40$), it is cMLP performs the best with NKDCD being the second best. At $F= 10, T=250, 500, 1000$, the AUROCs for the VAR with ulg- and hlg-lassos are both $86.1\pm 0.6$,  $86.2\pm 0.2$  and $93.0\pm 0.4$ respectively. The AUROCs for cMLP are reported to be $86.6 \pm 0.2$, $96.6\pm 0.2$ and $98.4\pm 0.1$ respectively,  while those for cLSTM are  $81.3  \pm 0.9$, $93.4 \pm 0.7$ and $ 96.0  \pm 0.1$ respectively. NKDCD with ilg-lasso penalty gives AUROC of $94.1\pm 0.7$, $98.5\pm 0.2$ and $99.8\pm 0.1$ while its counterpart with hlg-lasso is $94.1\pm 0.6$,  $98.5\pm 0.1$ and  $99.7\pm 0.1$ respectively.
When F is increased to $40$, for $T=250, 500, 1000$ the AUROCs for the VAR with ulg-lasso are  $70.0\pm 0.0$,  $80.6\pm 0.0$, and $84.7\pm0.0$, and with hlg-lasso  are $70.4\pm 0.0$, $81.1\pm 0.0$, and $84.7\pm0.0$ respectively. The AUROCs for cMLP are reported to be $84.0\pm 0.5$,  $89.6  \pm0.2$, and $95.5  \pm 0.3$, while those of cLSTM are $ 75.1 \pm 0.9 $, $87.8 \pm 0.4$, and $94.4 \pm0.5$ respectively.  NKDCD with ilg-lasso penalty gives AUROC of $80.8\pm 0.1$,  $92.0\pm 1.1$, and $98.7\pm 0.1$ while its counterpart with hlg-lasso is $82.6\pm 0.3$, $90.8\pm0.7$, and  $96.7\pm 0.2$ respectively.

\subsection{fMRI Data with Nonlinear Dynamics}
\begin{figure}[!htb]
    \centering
    \subfloat[\footnotesize Ground truth causal graph\label{2a}]{%
    \includegraphics[width=0.45\linewidth]{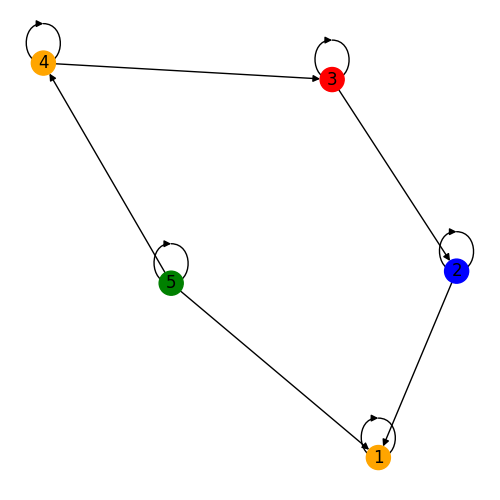}}
    \hfill
        \subfloat[\footnotesize Observed time series\label{2b}]{%
       \includegraphics[width=0.55\linewidth]{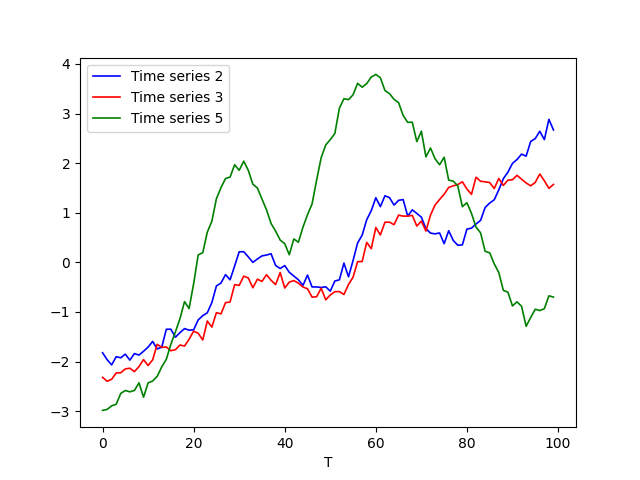}} 
    \caption{fMRI data}
    \label{fig:ogfmri}
    \end{figure}
Another benchmark, functional magnetic resonance imaging (fMRI), contains realistic, simulated Blood-oxygen-level dependent (BOLD) datasets for $28$ different underlying brain networks~\cite{smith2011network}. BOLD fMRI measures the neural activity of different regions of interest in the brain based on the changes in blood flow. Each region (i.e., a node in the brain network) has its associated time series of electrical excitation. All time series have an external input, white noise, and are fed through a nonlinear model~\cite {buxton1998dynamics}. All $28$ datasets are also made available on GitHub. Brain network $20$ has a dataset of size $2400 \times 5$ that corresponds to one of the $28$ brain networks. Its ground truth causal graph and three selected time series are depicted in Figure~\ref{fig:ogfmri}, in which time series $2$ and $3$ exhibit a causal relationship, while neither is causally linked to time series $5$.

\begin{table}[ht]
\centering
  \caption{fMRI Results}\label{tab:fmri}
  \begin{tabular}{||c| c c||} 
 \hline
  Dataset&\multicolumn{2}{|c|}{Brain network 20  }\\
  \hline
 Metrics &AUROC&AUPR\\\hline  \hline 
 NKDCD+Eq.~\ref{eq:glasso} &  $93.3\pm 0.0$ & $93.0\pm0.1$ \\
NKDCD+Eq.~\ref{eq:hlasso}  &$95.3\pm 0.0$ & $93.9\pm 0.1$  \\
VAR+Eq.~\ref{eq:ulgasso}  &$79.3\pm 3.9$ &  $82.5\pm 3.3$ \\
VAR+Eq.~\ref{eq:hlasso}  &$79.3\pm 3.9$ & $82.5\pm 3.3$ \\
 \hline
 \end{tabular}
\end{table}

The performance metrics are reported as the mean across five initializations with a $95\%$ confidence interval. The results, summarized in Table~\ref{tab:fmri}, show that our formulation is promising and can be utilized in realistic datasets. The AUROC and AUPR for the standard VAR method using ulg- and hlg-lasso penalties are $79.3\pm 3.9$ and  $82.5\pm 3.3$, respectively. Our method outperforms the VARs by having AUROC and AUPR of  $93.3\pm 0.0$ and $93.0\pm 0.1$ for the ilg-lasso penalty and $95.3\pm 0.0$ and $93.9\pm 0.1$ respectively for the hlg- counterpart. 

 \begin{table*}[t]
\centering
 \caption{DREAM3 Genetic Data}\label{tab:dream}
 \begin{tabular}{||c| c c | c c | c c |c c |c c ||} 
 \hline
 &\multicolumn{2}{|c|}{EC1 }& \multicolumn{2}{|c|}{EC2}&\multicolumn{2}{|c|}{Y1}&\multicolumn{2}{|c|}{Y2}&\multicolumn{2}{|c|}{Y3 }\\
  \hline
  Metric & AUROC &AUPR& AUROC &AUPR& AUROC &AUPR& AUROC &AUPR& AUROC &AUPR\\ \hline  \hline  
NKDCD+Eq.~\ref{eq:glasso} &  $81.1$ &$50.6$& $80.0$ &$45.1$ & $71.5$ & $45.5$& $65.2$ & $28.6$& $60.6$& $25.7$\\ 
NKDCD+Eq.~\ref{eq:hlasso}  &  $81.2$ &$50.6$& $80.2$ &$45.5$ & $71.6$ & $45.6$& $65.1$ & $29.2$& $60.4$& $25.6$\\ 
cMLP& $<65$&$<12$&$<65$ &$<14$  & $<70$&$<16$&$<60$&$<12$&$<60$&$<14$ \\
cLSTM & $<75$&$<12$&$<75$ &$<14$  & $<75$&$<12$&$<65$&$<16$&$<60$&$<16$ \\
VAR+Eq.~\ref{eq:ulgasso} &$51.8$& $2.6$&$50.9$ & $2.6$&   $50.2$& $2.9$& $ 51.3$& $5.3$&  $52.4$ &$7.6 $ \\
VAR+Eq.~\ref{eq:hlasso}   &$51.8$& $2.6$ &$50.9$ & $2.6$&   $50.2$& $2.9$& $ 51.3$& $5.3$&  $52.4$ &$7.6 $ \\
 \hline
 \end{tabular}  
\end{table*}  

\subsection{Dream3 Genetic Data with Nonlinear Dynamics}  
\begin{figure}[!htb]
    \centering
    \subfloat[\footnotesize Ground truth causal graph \label{1a}]{%
    \includegraphics[width=0.45\linewidth]{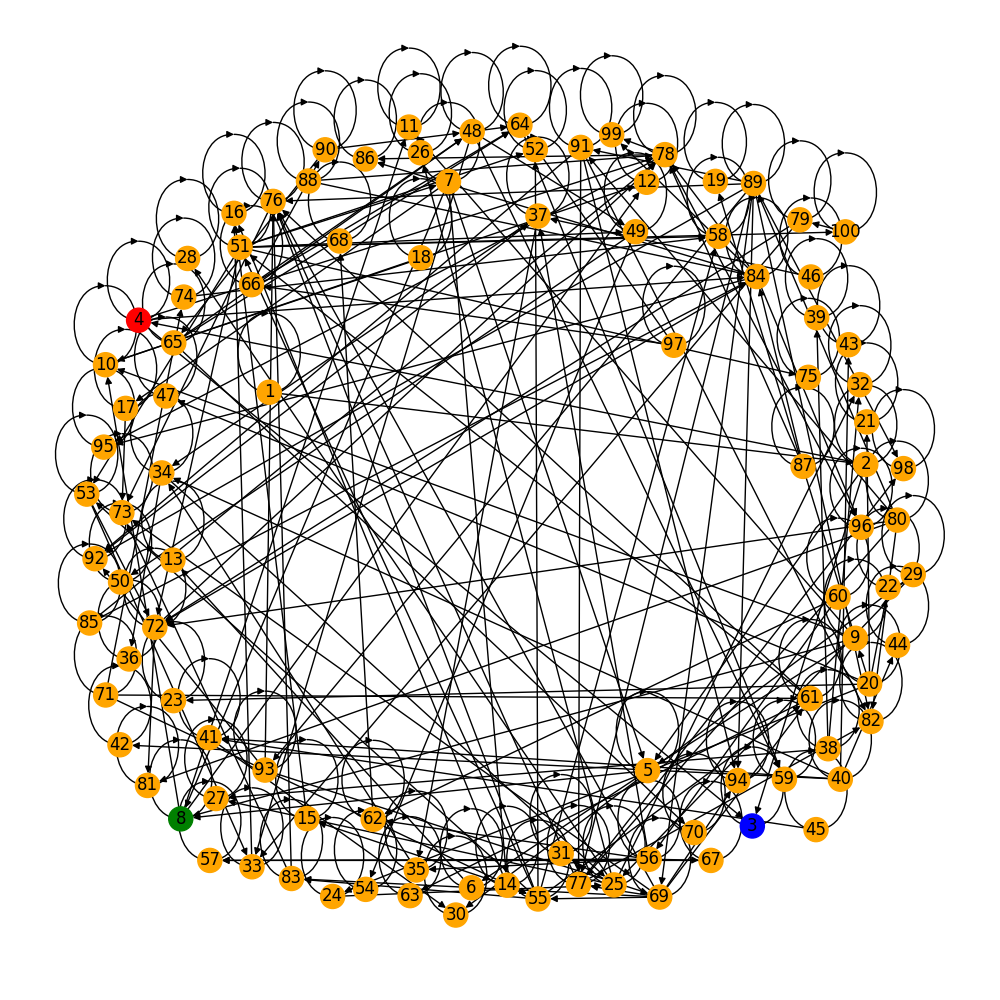}}
    \hfill
        \subfloat[\footnotesize Observed time series (selected)\label{1b}]{
       \includegraphics[width=0.53\linewidth]{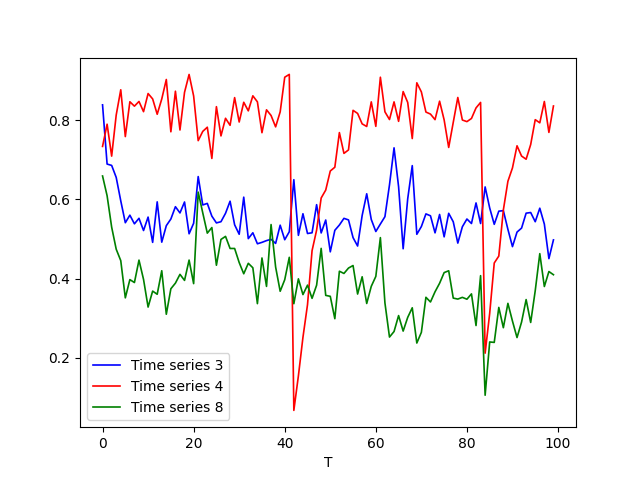}} 
    \caption{Dream3 genetic data for Yeast 1}
    \label{fig:ogy1}
    \end{figure}
\begin{figure}[ht]
\subfloat[\footnotesize  AUROC bar-plots \label{fig:dreamauroc}]{ 
    \includegraphics[width=0.49\linewidth]{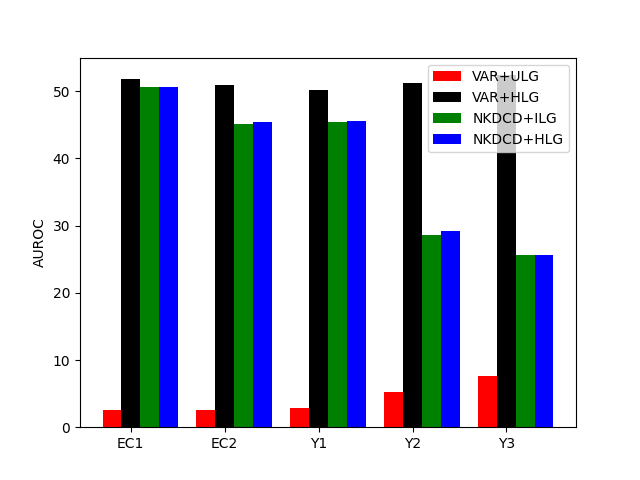}}
        \hfill
\subfloat[\footnotesize  AUPR bar-plots 
\label{fig:dreamAUPR}]{ 
        \includegraphics[width=0.49\linewidth]{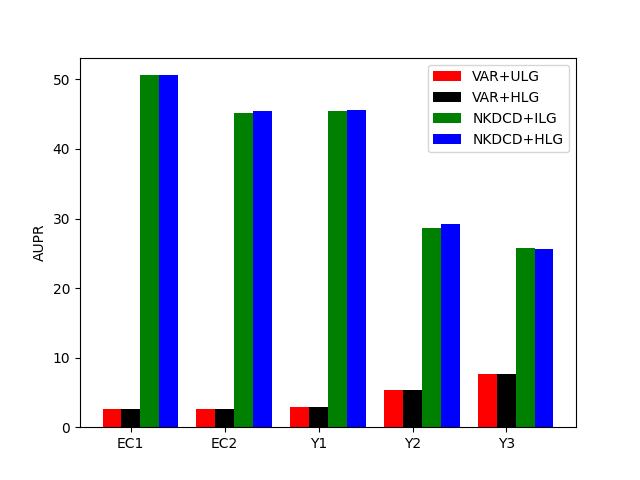}}
    \caption{Accuracy plots of DREAM3 dataset }
    \end{figure}
\begin{figure}[ht]
    \centering
    \subfloat[\footnotesize EC1\label{dreamec1}]{ \includegraphics[width=0.4\linewidth]{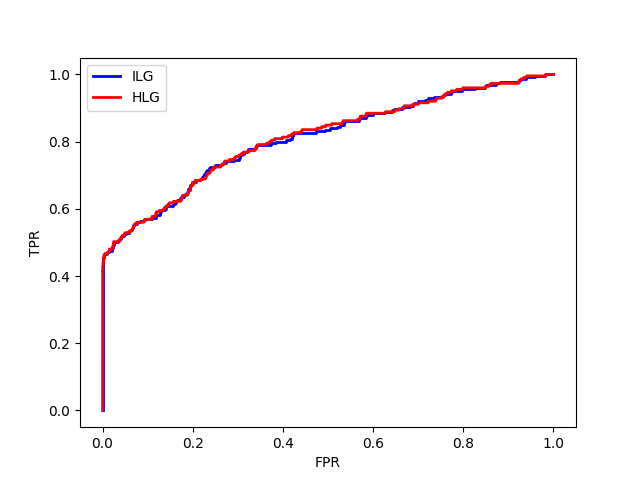}}
    \hfill
   \subfloat[\footnotesize EC2\label{dreamec2}]{
\includegraphics[width=0.4\linewidth]{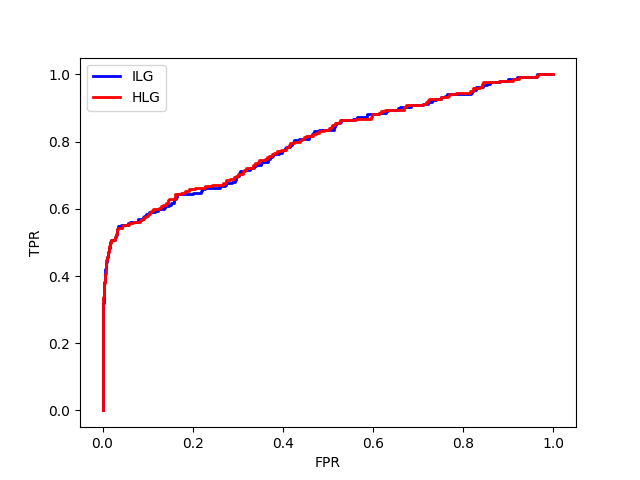}} 
   \hfill
   \subfloat[\footnotesize Y1\label{dreamy1}]{
\includegraphics[width=0.4\linewidth]{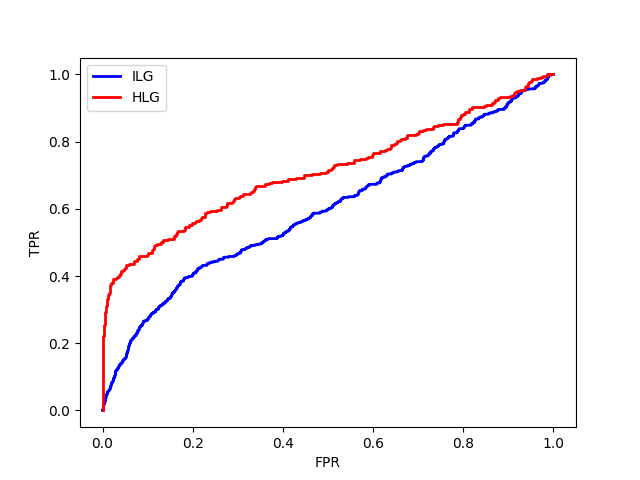}} 
   \hfill
   \subfloat[\footnotesize Y2\label{dreamy2}]{
\includegraphics[width=0.4\linewidth]{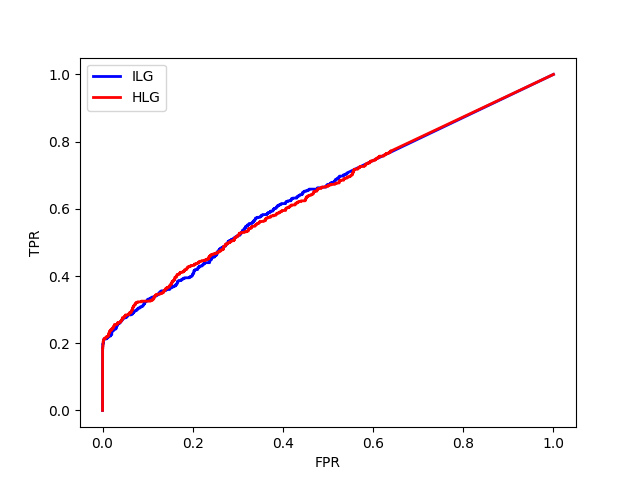}} 
   \hfill
   \subfloat[\footnotesize Y3\label{dreamy3}]{
\includegraphics[width=0.4\linewidth]{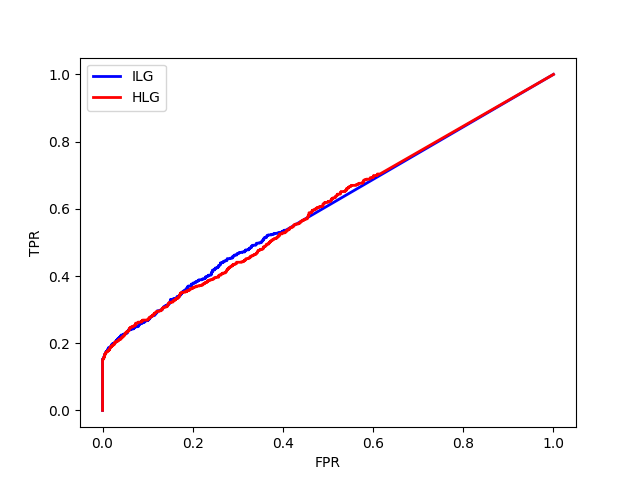}} 

    \caption{ROC plots of DREAM3 dataset}
    \label{fig:dreamrocog}
    \end{figure}

DREAM3 challenge dataset \cite{prill2010towards} provides a rigorous benchmark for comparing GC detection methods for gene regulatory networks~\cite{lim2015operator, lebre2009inferring, tank2021neural}. The data are synthetically generated to mimic continuous gene expression and regulation dynamics, incorporating multiple unobserved hidden factors. The DREAM3 challenge comprises $5$ distinct simulated datasets, each associated with different ground truth GC dependencies. Specifically, there are two datasets based on E.coli (EC 1-2) and three datasets based on Yeast (Y 1-3). In each dataset, there are $100$ time series, with $46$ replicates sampled at $21$ time points, resulting in a times series data of size $100 \times 966$. This setup presents a significant challenge due to the limited amounts of data relative to the complexity of the networks and underlying dynamic interactions, as shown in Figure~\ref{1a}, whereas $3$ selected time series elements from a single replicate of the Y1 dataset are shown in Figure~\ref{1b}. This snapshot provides insight into the intricacies of the gene expression dynamics encapsulated within the dataset, offering a stage for our GC estimation efforts. In this simulation, the ADAM~\cite{kingma2014adam} with parameters $\beta =  (0.9, 0.999)$ and $\epsilon = 10^{-8}$, was used to initially update the learnable parameters in place of the ordinary gradient descent, followed by the proximal gradient descent was used to sparsify the lag matrices $W_l,l\in[1,L]$. We applied our results to all five data sets. Results for AUROC are visualized in Figure~\ref{fig:dreamauroc} and for AUPR in Figure~\ref{fig:dreamAUPR}. The ROC plots for both the ilg- and hlg-lassos are shown in Figure~\ref{fig:dreamrocog}. 

We compared our results based on the bar plots in the recent work reported in~\cite{tank2021neural} in Table~\ref{tab:dream}. For the EC1 dataset, the AUROC values for the standrad VAR with ulg-lasso and with hlg-lasso, cMLP, cLSTM, NKDCD with ilg-lasso, and with hlg-lasso respectively are $51.8,51.8$,$65$, $75$, $81.1$, $81.2$ and their corresponding AUPRs are $2.6$, $2.6$, $12$, $12$, $50.6$, $50.6$, respectively. Following the same pattern for the EC2 dataset, the AUROCs are $50.9$, $50.9$,$65$, $75$, $80.0$, $80.2$ and their corresponding AUPRs are $2.6$, $2.6$, $14$, $14$, $45.1$, $41.5$, respectively.
The performance of the Y1 dataset for AUROCs are $50.2$, $50.2$,$70$, $75$, $71.5$, $71.6$ and their corresponding AUPRs are $2.9$, $2.9$, $16$, $12$, $45.5$, $45.6$, respectively.  The Y2 dataset yields AUROCs are as, $51.3$, $51.3$,$60$, $65$, $65.2$, $65.1$ and their corresponding AUPRs as, $5.3$, $5.3$, $12$, $16$, $28.6$, $29.2$, respectively. The Y3 Dataset gives AUROCs as, $52.4$, $52.4$,$60$, $60$, $60.6$, $60.4$ and their corresponding AUPRs as, $7.6$, $7.6$, $<14$, $<16$, $25.7$, $25.6$, respectively. Results suggest that the dynamics for Y 1-3 may be more complex than that for EC 1-2. In all cases, our results outperform the existing linear and nonlinear methods.

\section{Conclusion}
We have introduced the NeuroKoopman Dynamic Causal Discovery (NKDCD) framework, which leverages a Koopman-inspired regularized deep neural network autoencoder architecture for the inference of nonlinear Granger causality and underlying dynamics in observed multivariate time series data. This framework enables the learning of underlying dynamics and their interdependencies through sparse nonlinear regressive models. By employing a data-driven NN-based basis function for lifting the NAR dependency for a linear Koopman embedding and next for learning a linear vector autoregressive model in the lifted domain, we infer Granger causality and the underlying dynamics. We employ structured group-level sparsity-inducing penalties that preserve the structure of interdependencies of the original variables, ensuring reliable model learning. Validation through simulations on various application datasets shows that our NKDCD outperforms the state-of-the-art results from recent literature. 

\IEEEpeerreviewmaketitle
\bibliographystyle{IEEEtran}
 \bibliography{cd}

 \vskip -1.8\baselineskip plus -1fil
\begin{IEEEbiography}[{\includegraphics[width=1in,height=1.2in,clip,keepaspectratio]{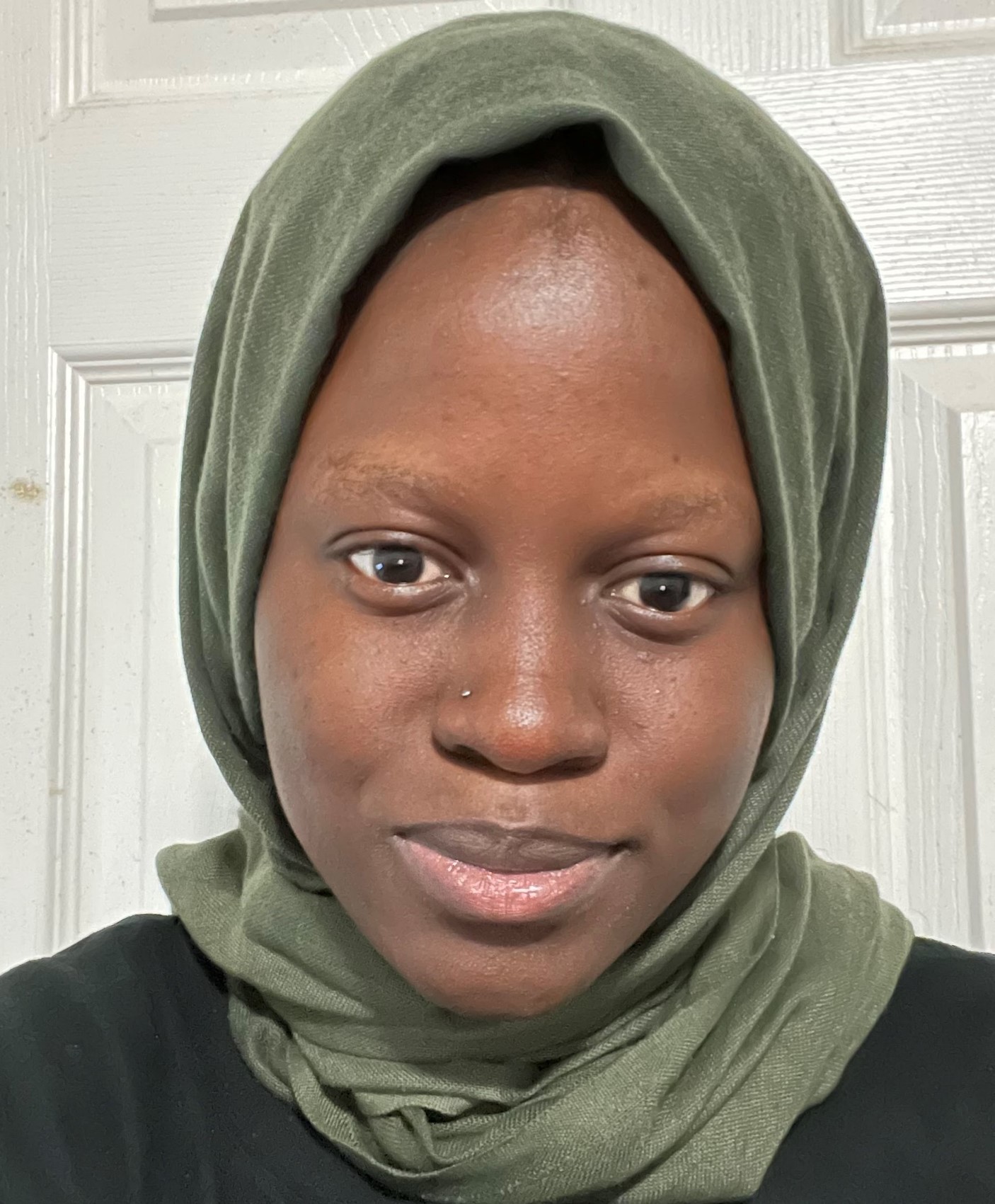}}]{Rahmat Adesunkanmi} is a Ph.D. student at the Department of Electrical and Computer Engineering, Iowa State University, Ames, USA.  She received her B.S. degree from the University of Ibadan, Nigeria, in 2015 and subsequently her master's degree from Iowa State University. Her research interests include noise-robust Machine learning techniques and Time-series analysis. She is an active member of the Graduate Society for Women Engineers.
\end{IEEEbiography}

\vskip -1.0\baselineskip plus -1fil
\begin{IEEEbiographynophoto}
{\bf Balaji Sesha Srikanth Pokuri} received the B.Tech degree in electrical engineering from Indian Institute of Technology, Roorkee, India, in 2017. He is currently working toward the Ph.D. degree in electrical engineering with the Department of Electrical and Computer Engineering, Iowa State University, Ames, IA, USA.

He worked as a Quantitative Analyst in eFX Trading business of HSBC Global Markets, from 2017 to 2021. His current research includes data-driven modeling, control, optimization, and artificial-intelligence-based approaches for precision agriculture.
\end{IEEEbiographynophoto}

\vskip -1.0\baselineskip plus -1fil
\begin{IEEEbiography}[{\includegraphics[width=1in,height=1.2in,clip,keepaspectratio]{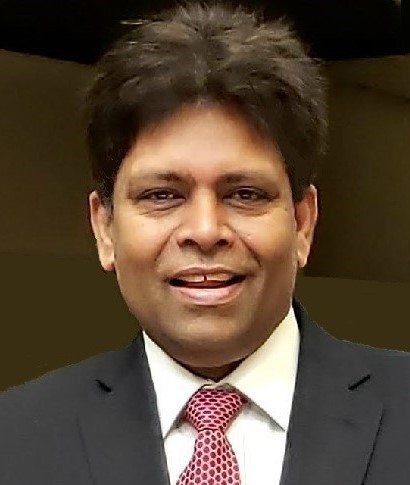}}]{Ratnesh Kumar} is a Palmer Professor at the Department of Electrical and Computer Engineering, Iowa State University, where he directs the {\bf ESSeNCE} ({\bf E}mbedded {\bf S}oftware, {\bf Se}nsors, {\bf N}etworks, {\bf C}yberphysical, and {\bf E}nergy) Lab. 
Previously, he held a faculty position at the University of Kentucky and various visiting positions with the University of Maryland (College Park), the Applied Research Laboratory at the Pennsylvania State University (State College), NASA Ames, the Idaho National Laboratory, the United Technologies Research Center, and the Air Force Research Laboratory. 
He received a B. Tech. degree in Electrical Engineering from IIT Kanpur, India, in 1987 and the M.S. and Ph.D. degrees in Electrical and Computer Engineering from the University of Texas at Austin in 1989 and 1991 respectively. Ratnesh is a Fellow of IEEE, also a Fellow of AAAS, and was a Distinguished Lecturer of the IEEE Control Systems Society. 
He is a recipient of {\em D. R. Boylan Eminent Faculty Award for Research} and {\em Award for Outstanding Achievement in Research} from Iowa State University, and also the {\em Distinguished Alumni Award} from IIT Kanpur. Ratnesh received Gold Medals for the Best EE Undergrad, the Best All Rounder, and the Best EE Project from IIT Kanpur, and the Best Dissertation Award from UT Austin, the Best Paper Award from the IEEE Transactions on Automation Science and Engineering, and has been Keynote Speaker and paper award recipient from multiple conferences. He is or has been an editor of several journals (including of IEEE, SIAM, ACM, Springer, IET, MDPI).
\end{IEEEbiography}
 
\end{document}